\documentclass[11pt]{article}
\pdfoutput=1

\usepackage{authblk} 
\usepackage{acl2023}

\usepackage{times}
\usepackage{latexsym}
\usepackage{amsfonts}
\usepackage{arydshln}

\usepackage[T1]{fontenc}
\usepackage[utf8]{inputenc}

\usepackage{microtype}

\usepackage{inconsolata}

\usepackage{tipa}
\usepackage{booktabs}
\usepackage{multirow}
\usepackage{makecell}
\usepackage{threeparttable}

\usepackage{microtype}

\usepackage{inconsolata}
 
\usepackage{xcolor,colortbl}
\usepackage{color,soul}
\usepackage{adjustbox}
\usepackage{amsmath,amssymb}
\usepackage{graphicx}

\usepackage[inline]{enumitem}

\def\bng{\bngx}

\font\bngx=bang10

\def\*#1*#2{o\null{#2}{#1}}

\def\sh#1{\setbox0=\hbox{#1}%
     \kern-.02em\copy0\kern-\wd0
     \kern.04em\copy0\kern-\wd0
     \kern-.02em\raise.0433em\box0 }

\title{AxomiyaBERTa: A Phonologically-aware Transformer Model for Assamese}

\author[]{\rm {\bf Abhijnan Nath}}
\author[]{\rm {\bf Sheikh Mannan}}
\author[]{\rm {\bf Nikhil Krishnaswamy}}
\affil[]{Situated Grounding and Natural Language (SIGNAL) Lab \authorcr Department of Computer Science, Colorado State University \authorcr Fort Collins, CO, USA \authorcr {\tt \{abhijnan.nath,sheikh.mannan,nkrishna\}@colostate.edu}}

\begin{document}
\maketitle
\begin{abstract}
Despite their successes in NLP, Transformer-based language models still require extensive computing resources and suffer in low-resource or low-compute settings.  In this paper, we present AxomiyaBERTa, a novel BERT model for Assamese, a morphologically-rich low-resource language (LRL) of Eastern India. AxomiyaBERTa is trained only on the masked language modeling (MLM) task, without the typical additional next sentence prediction (NSP) objective, and our results show that in resource-scarce settings for very low-resource languages like Assamese, MLM alone can be successfully leveraged for a range of tasks.  AxomiyaBERTa achieves SOTA on token-level tasks like Named Entity Recognition and also performs well on ``longer-context'' tasks like Cloze-style QA and Wiki Title Prediction, with the assistance of a novel embedding disperser and phonological signals respectively. Moreover, we show that AxomiyaBERTa can leverage phonological signals for even more challenging tasks, such as a novel cross-document coreference task on a translated version of the ECB+ corpus, where we present a new SOTA result for an LRL. Our source code and evaluation scripts may be found at \url{https://github.com/csu-signal/axomiyaberta}.
\end{abstract}

\vspace*{-2mm}
\section{Introduction}
\label{sec:intro}
\vspace*{-2mm}

Transformer-based neural architectures such as BERT~\cite{devlin-etal-2019-bert} have revolutionized natural language processing (NLP).  The ability to generate contextualized embeddings that both preserve polysemous word sense and similarity across dimensions through self-attention has contributed to significant improvements in various NLP tasks~\cite{ethayarajh-2019-contextual}. Despite their successes, Transformers come at a high computational cost~\cite{zhao-etal-2022-fine} and still suffer from long-standing issues pertaining to data-hunger and availability of training resources.  One effect of the dependency on big data is the continued proliferation of sophisticated NLP for well-resourced languages while low-resourced languages (LRLs) continue to be underrepresented, and the disparities continue to grow~\cite{joshi-etal-2020-state}. 

This is particularly true for languages of India and South Asia where English is widely spoken among the educated and urban population. Therefore, those in India most likely to use and develop NLP may freely do so in English, but sole speakers of local Indian languages may remain effectively isolated from human language technology in their native tongues. While strides have been made in NLP for widely-spoken Indian languages (e.g., Hindi, Bengali, Marathi, Tamil, etc.), India is home to about a thousand languages, over 100 of which are considered ``major''\footnote{\url{https://censusindia.gov.in/census.website/data/census-tables}} but are not widely represented in NLP research.  This lack of representation also precludes insights from those languages from contributing to the field~\cite{bender2019benderrule}.

In this paper, we present {\bf AxomiyaBERTa}, a novel Transformer language model for the Assamese language.\footnote{Despite the name, AxomiyaBERTa is an ALBERT variant, not a RoBERTa variant. The name is derived from {\it Axomiya} ({\bng Asmiiya}, \textipa{/OxOmija/}), the native term for the Assamese language, plus ``BERTa'' from both BERT and {\it barta}, meaning ``conversation.''  The name also recalls {\it Asom Barta}, the official newsletter of the Government of Assam.} AxomiyaBERTa has been trained in a low-resource and limited-compute setting, using only the masked language modeling (MLM) objective. Beyond a model for a new language, our novel contributions are as follows:

\begin{itemize}
    \vspace*{-1mm}
    \item Use of a novel combined loss technique to disperse AxomiyaBERTa's embeddings;
    \vspace*{-2mm}
    \item Addition of phonological articulatory features as an alternate performance improvement in the face of omitting the NSP training objective for longer-context tasks;
    \vspace*{-2mm}
    \item Evaluation on event coreference resolution, which is novel for Assamese.
\end{itemize}

AxomiyaBERTa achieves competitive or state of the art results on multiple tasks, and demonstrates the utility of our approach for building new language models in resource-constrained settings.

%\textipa{aI p\super hi: eI}

%{\beng অসমীয়া}

\vspace*{-2mm}
\section{Related Work}
\label{sec:related}
\vspace*{-2mm}

%Most work on Transformer-based language models has been performed on high-resource languages, English in particular.
Multilingual large language models (MLLMs) trained over large Internet-sourced data, such as MBERT and XLM~\cite{conneau-etal-2020-unsupervised}, provide resources for approximately 100 languages, many of which are otherwise under-resourced in NLP.  However, multiple publications~\cite{virtanen2019multilingual,scheible2020gottbert,tanvir2021estbert} have demonstrated that multilingual language models tend to underperform monolingual language models on common tasks; the ``multilingual'' quality of MLLMs may not be enough to assure performance on LRL tasks, due to language-specific phenomena not captured in the MLLM.

Since languages that share recent ancestry or a {\it Sprachbund} tend to share features, there has also been development of models and resources for languages from distinct regions of the world.  South Asia is one such ``language area,'' where even unrelated languages may share features (e.g., 4-way voice/aspiration distinctions, SOV word order, retroflex consonants,  heavy use of light verbs).  As such, researchers have developed region-specific models for South Asian languages such as IndicBERT~\cite{kakwani-etal-2020-indicnlpsuite} (11 languages, 8.8 billion tokens) and MuRIL~\cite{DBLP:journals/corr/abs-2103-10730} (17 languages, 16 billion tokens).

Subword tokenization techniques like byte-pair encoding (BPE)~\cite{sennrich-etal-2016-neural} yield comparatively better performance on LRLs by not biasing the vocabulary toward the most common words in a specific language, but BPE tokens also further obscure morphological information not immediately apparent in the surface form of the word.  \citet{nzeyimana-niyongabo-rubungo-2022-kinyabert} tackle this problem for Kinyarwanda using a morphological analyzer to help generate subwords that better capture individual morphemes.  However, despite similar morphological richness of many Indian languages, and likely due to similar reasons as outlined above, the dearth of NLP technology for most Indian languages extends to a lack of morphological parsers. We hypothesize that adding phonological features can also capture correlations between overlapping morphemes.

Previous NLP work in Assamese includes studies in corpus building~\cite{sarma-etal-2012-structured,laskar-etal-2020-enascorp1,pathak-etal-2022-asner}, POS tagging~\cite{kumar-bora-2018-part}, WordNet~\cite{bharali2014analytical,sarmah2019development} structured representations~\cite{sarma2012structured}, image captioning~\cite{nath-etal-2022-image}, and cognate detection \cite{nath-etal-2022-phonetic}.
There does not exist, to our knowledge, significant work on Assamese distributional semantics, or any monolingual, Transformer-based language model for the Assamese language evaluated on multiple tasks.
%... other RW ...

Our work complements these previous lines of research with a novel language model for Assamese, which further develops an initial model first used in \citet{nath-etal-2022-phonetic}. We account for the lack of an Assamese morphological analyzer with additional phonological features and task formulations that allow for strategic optimization of the embedding space before the classification layer.

\vspace*{-2mm}
\subsection{Assamese}
\label{ssec:assamese}
\vspace*{-1mm}

Assamese is an Eastern Indo-Aryan language with a speaker base centered in the Indian state of Assam.  It bears similarities to Bengali and is spoken by 15 million L1 speakers (up to 23 million total speakers).  Its literature dates back to the 13th c. CE. It has been written in its modern form since 1813, is one of 22 official languages of the Republic of India, and serves as a {\it lingua franca} of the Northeast Indian region~\cite{jain2004indo}.

Despite this, Assamese data in NLP resources tends to be orders of magnitude smaller than data in other languages, even in South Asian region-specific resources (see Table~\ref{tab:datasizes}).

\begin{table}[h!]
    \vspace*{-2mm}
    \centering
    \begin{tabular}{lllll}
        \toprule
         & \small{\textit{\textbf{as}}} & \small{\textit{\textbf{bn}}} & \small{\textit{\textbf{hi}}} & \small{\textit{\textbf{en}}} \\
        \cmidrule(lr){1-5}
        \small{\bf CC-100} & \small 5 & \small 525 & \small 1,715 & \small 55,608 \\
        \small{\bf IndicCorp} & \small 32.6 & \small 836 & \small 1,860 & \small 1,220 \\
        \bottomrule
    \end{tabular}
    \vspace*{-2mm}
    \caption{CC-100~\cite{conneau-etal-2020-unsupervised} and IndicCorp~\cite{kakwani-etal-2020-indicnlpsuite} data sizes (in millions of tokens) for Assamese, Bengali, Hindi, and English.}
    \label{tab:datasizes}
    \vspace*{-2mm}
\end{table}

Assamese bears a similar level of morphological richness to other Indo-Aryan and South Asian languages, with 8 grammatical cases and a complex verbal morphology. Despite these points of comparison, Assamese has some unique phonological features among Indo-Aryan languages, such as the use of alveolar stops \textipa{/t\super{(h)}/}, \textipa{/d\super{(H)}/}, velar fricative \textipa{/x/}, and approximant \textipa{/\textturnr/}. This atypical sound pattern motivates the use of phonological signals in our model. Moreover, both the pretraining and task-specific corpora we use contain a large proportion of loanwords (e.g., from English) or words cognate with words in higher-resourced languages (e.g., Bengali). These words rendered with Assamese's unique sound pattern result in distinct, information-rich phoneme sequences.

\vspace*{-2mm}
\section{Methodology}
\label{sec:method}
\vspace*{-2mm}

\subsection{Pretraining}
\label{ssec:pretraining}
\vspace*{-1mm}

We trained on four publicly-available Assamese datasets: Assamese Wikidumps\footnote{\url{https://archive.org/details/aswiki-20220120}}, OSCAR~ \cite{suarez2019asynchronous}\footnote{\url{https://oscar-corpus.com}}, PMIndia~ \cite{haddow2020pmindia}\footnote{\url{https://paperswithcode.com/dataset/pmindia}}, the Common Crawl (CC-100) Assamese corpus~ \cite{conneau-etal-2020-unsupervised}\footnote{\url{https://paperswithcode.com/dataset/cc100}}, as well as a version of the ECB+ Corpus~\cite{cybulska-vossen-2014-using} translated to Assamese using Microsoft Azure Translator. In total, after preprocessing, the training data amounts to approximately 26 million space-separated Assamese tokens.\footnote{In resource-scarce settings, especially for LRLs, it is challenging to find large monolingual corpora for Transformer training. For instance, XLM-R was pretrained on about 164B tokens, of which only 5M were Assamese (see Table~\ref{tab:datasizes}). However \citet{ogueji-etal-2021-small} suggest that for LRLs, smaller datasets can actually work {\it better} than joint training with high-resourced parallel corpora.}

AxomiyaBERTa (66M parameters) was trained as a ``light'' ALBERT (specifically {\tt albert-base-v2}) \cite{lan2019albert} model with {\it only} the MLM objective \cite{devlin-etal-2019-bert}, and no next sentence prediction (NSP), for 40 epochs (485,520 steps) with a vocabulary size of 32,000 and a SentencePiece BPE tokenizer \cite{kudo-richardson-2018-sentencepiece}. Tokenization methods like BPE can obfuscate certain morphological information. However, without a publicly-available morphological analyzer for Assamese, our motivation was to examine if phonological correlations might pick up similar information across different tasks while keeping model architecture and tokenizer consistent. We trained on 1 NVIDIA A100 80 GB device with a batch size of 32 and a sequence length of 128 for approximately 72 hours. Table~\ref{tab:axberta-config} in Appendix~\ref{app:train-config} shows all specific pretraining configuration settings.

% Here we explore the extent to which this holds for Assamese with AxomiyaBERTa's pretraining regime.

\vspace*{-1mm}
\subsubsection{Special Token Vocabulary}
\label{sssec:special-tokens}

The AxomiyaBERTa vocabulary includes two special trigger tokens: {\tt <m>} and {\tt </m>}.  These act as separators {\it a la} the BERT {\tt [SEP]} token, meaning that contextualized representations of these tokens were trained into the AxomiyaBERTa embedding space. Prior to pretraining, the translated ECB+ Corpus was annotated with these tokens surrounding event mentions. Since AxomiyaBERTa was not trained using the next sentence prediction objective (see Sec.~\ref{sssec:articulatory}), its embedding space needs those special triggers as separators between segments instead of the {\tt [SEP]} tokens that segregate the token type IDs.

\vspace*{-1mm}
\subsection{Fine-tuning}
\vspace*{-1mm}

AxomiyaBERTa pretraining created a task-agnostic model optimized for the grammar and structure of Assamese. This model was then fine-tuned to achieve good performance on a number of different tasks. Beyond the task-specific fine-tuning, we made use of two auxiliary techniques: an {\it embedding disperser}, that optimized the AxomiyaBERTa embedding space away from severe anisotropy, and {\it phonological or articulatory attention} that acted as a single-head attention layer attending to both token-level and candidate-option level phonological signals. We first discuss these two techniques, followed by the specific task formulations we evaluated on. Note that the embedding disperser was used at the fine-tuning stage for Cloze-QA \textit{only} due to severe anisotropy of the embedding space (Fig.~\ref{fig:cosine_embed_small} and Fig.~\ref{fig:cosine_embed}, Appendix \ref{app:emb-disp}).

\vspace*{-1mm}
\subsubsection{Embedding Disperser}
\label{sssec:disperser}

\begin{figure}
  \centering
  \includegraphics[width=0.48\textwidth]{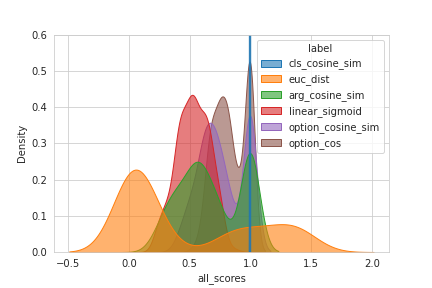}
  \vspace*{-2mm}
\caption{Kernel Density Estimation plots for within-set features of each component of the phonologically-aware embedding disperser (Fig.~\ref{fig:emb_disperser}). {\tt option\_cos} is the output of the auxiliary discriminator while {\tt linear\_sigmoid} represents the linear layer and {\tt euc\_dist} represents L$^2$ norm between the raw {\tt [CLS]} token embeddings. See Fig.~\ref{fig:cosine_embed} in Appendix~\ref{app:emb-disp} for equivalent ``beyond-set'' plots.} 
\label{fig:cosine_embed_small}
\vspace*{-4mm}
\end{figure}

\begin{figure*} 
    \centering
    \includegraphics[scale=1.9,width=\textwidth]{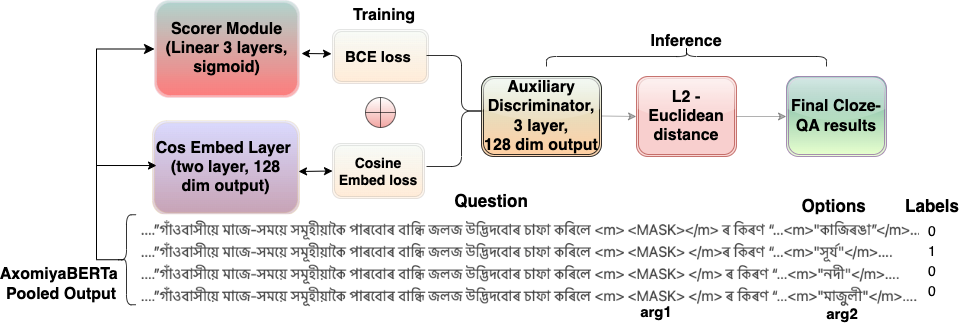} 
    \vspace*{-2mm}
    \caption{Embedding disperser architecture with Cosine Embedding and Binary Cross Entropy (BCE) Loss. Cloze-QA data used as example.}
    \label{fig:emb_disperser}
    \vspace*{-4mm}
\end{figure*}

% \begin{figure*} 
%     \centering
%     \includegraphics[scale=1.9,width=\textwidth]{plots/KDEplot_truesamples_pan.png} 
%     \vspace*{-2mm}
%     \caption{Kernel Density Estimation plots from various features sets of the Embedding Disperser model.  }
%     \label{fig:cosine_embed_small}
%     \vspace*{-4mm}
% \end{figure*}

Without a meaningful objective to force embedding vectors apart during training, they trend toward an arbitrary center in $\mathbb{R}^d$ space.  This phenomenon has also been observed by \citet{gao2018representation}, \citet{ethayarajh-2019-contextual}, and \citet{demeter-etal-2020-stolen}, among others. In \citet{nath-etal-2022-phonetic}, evidence was presented that the effect is more pronounced in smaller models. An effect of this can be illustrated by embeddings from an example task, Cloze-style question answering (Cloze-QA):

Let a ``set'' of embeddings consist of representations for a question (or context) $Q$ and associated candidate answers $\{A,B,C,D\}$. ``Within-set'' cosine similarities represent the cosine similarities between $(Q+i,Q+j)$ for each candidate answer $i \in \{A,B,C,D\}$ and each other candidate $j \in \{A,B,C,D\} \text{ \textit{where} } i \neq j$. ``Beyond-set'' cosine similarities represent similarities between all pairs in a candidate-plus-answers set compared to other such embedding sets from different questions. Fig.~\ref{fig:cosine_embed_small} shows KDE plots for various similarity metrics taken ``within-set'' for a random sample of 100 sets from the Cloze-QA dev set (see Sec.~\ref{sssec:cloze-qa} for more details on the data).  The blue spike at 1 for {\tt cls\_cosine\_sim} shows how similar all {\tt[CLS]} token embeddings are to each other, given AxomiyaBERTa's extremely anisotropic embedding space after pretraining. This makes it difficult to optimize a classification boundary during fine-tuning using standard techniques.

% [description of technique - what is being done here to disperse the embeddings?]

Therefore, to disperse the embedding space for greater discriminatory power, we used a combination of Binary Cross Entropy loss and Cosine Embedding loss to train the model. The architecture is shown in Fig.~\ref{fig:emb_disperser}. The key components are:
\begin{enumerate*}[label=\roman*)]
    \item a {\it cosine embedding layer} that takes in {\tt arg1} (context) and {\tt arg2} (candidate) representations along with a {\tt [CLS]} representation and outputs a 128D embedding into the cosine embedding loss function, and
    \item an {\it auxiliary discriminator} that considers only {\tt arg2} and {\tt [CLS]} representations.
\end{enumerate*}

Mathematically, 

\vspace*{-4mm}
{\small
$$
L_{B C E}=-\frac{1}{n} \sum_{i=1}^n\left(Y_i \cdot \log \hat{Y}_i+\left(1-Y_i\right) \cdot \log \left(1-\hat{Y}_i\right)\right)
$$}
\vspace*{-6mm}

{\small
$$
L_{COS}(x, y)= \begin{cases}1-\cos \left(x_1, x_2\right), & \text { if } y=1 \\ \max \left(0, \cos \left(x_1, x_2\right)-\operatorname{m}\right), & \text { if } y=-1\end{cases}
$$}
\vspace*{-6mm}

{\small
$$
L_{COMB} = \alpha L_{B C E} + L_{COS}(x, y)
$$
}
\vspace*{-2mm}

where $\operatorname{m}$ represents the margin for the cosine loss and $\alpha$ is 0.01. $x_1$ corresponds to {\tt arg1} and $x_2$ corresponds to {\tt arg2}. $y = 1$ if $x_2$ is the correct answer and $y = -1$ if not. At inference, we computed Euclidean distance between the embedding outputs of the auxiliary discriminator and the cosine embedding layer with a threshold $T$ of 4.45, found through hyperparameter search.

{\tt option\_cosine\_sim} in Fig.~\ref{fig:cosine_embed_small} shows the outputs of the embedding disperser's cosine embedding layer while {\tt option\_cos} shows the outputs of the auxiliary discriminator.  In both cases we see distinct distributions that separate correct and incorrect answers.  Works such as \citet{cai2021isotropy} present evidence of such cases of global token anisotropy in other Transformer models and suggest that creating such local isotropic spaces leads to better results in downstream tasks.

\vspace*{-1mm}
\subsubsection{Phonological/Articulatory Attention}
\label{sssec:articulatory}

While the NSP objective is effective at training LLMs to encode long-range semantic coherence \cite{shi-demberg-2019-next}, it comes at a significant additional computational cost.  Moreover, for very low-resource languages like Assamese, a lack of available long document or paragraph data means there may not exist a sufficient volume of coherent consecutive sentences in the training data. %For instance, AxomiyaBERTa's training corpora consist of many more independent single sentences than longer documents that could be used to generate positive samples for such a binary classification objective.   

% We hypothesize that smaller models like AxomiyaBERTa 

We hypothesize that when fine-tuning a smaller model like AxomiyaBERTa in a resource-constrained setting, adding phonological signals to the latent representations of text samples allows us to achieve a balanced trade-off between possible information loss due to reduced supervision (no NSP objective) and improved task-specific performance, at a lower compute cost.

% during task fine-tuningsmaller models like AxomiyaBERTa 

% can nonetheless compensate for the information lost without such a training objective by using additional phonological signals during task fine-tuning. 

%  Adding phonological features does not directly make up performance lost due to omitting NSP, but increases performance elsewhere by correlating phonologically similar forms (e.g., morphological derivations that otherwise have dissimilar embeddings). We’ll rephrase this in the introduction, rephrase the relevant contributions bullet point to make this clearer, and edit the discussion in Sec. 3.2.2 to reflect this framing (e.g., “compensate for the information lost…” -> “improve performance elsewhere…”).

Previous works (e.g., \citet{mortensen-etal-2016-panphon,rijhwani2019zero,nath-etal-2022-generalized}) have shown that phonological features are useful for both token-level ``short-context'' tasks like NER or loanword detection as well as ``longer-context'' tasks like entity linking. We fine-tune for longer-context tasks by encoding candidate answers as phonological features and the pooled embedding of the context, and computing the relative difference in mutual information between each candidate answer and the context. High variance in cosine similarities within pairs in a context-candidate set is due to the phonological signals. Table~\ref{tab:mean_variance_pan_qa} shows that the mean, standard deviation, and variances of {\tt[CLS]} token cosine similarities for pretrained AxomiyaBERTa are much smaller than those extracted from XLM, but fine-tuning with phonological signals brings AxomiyaBERTa's values much closer XLM's.

% Previous works (e.g., \citet{mortensen2016panphon,rijhwani2019zero,nath-etal-2022-generalized}) have shown that information-rich articulatory features can be successfully leveraged for both token-level ``short-context'' tasks like NER or loanword detection as well as ``longer-context'' tasks like entity linking. The embedding space of
% %smaller models can be anisotropic as evidenced by \cite{nath-etal-2022-phonetic}
% a model can also be anisotropic, as evidenced by \citet{ethayarajh-2019-contextual} and \citet{demeter-etal-2020-stolen}. This means that the individual embeddings, while sufficiently encoding the ``candidates'' preceded by the ``context,'' have less absolute variance in cosine similarity among themselves within a QA set and between them and other QA sets as well as when compared to those from larger models like XLM. For instance, Table~\ref{tab:mean_variance_pan_qa} shows that both the standard deviation and variances of {\tt[CLS]} token cosine similarities for AxomiyaBERTa are much smaller than that of XLM while adding articulatory signals increasing both values in case of the former. 

\begin{table}[h!]
    \vspace*{-1mm}
    \centering
    \begin{tabular}{llll}
        \toprule
        
         & \small{\textbf{AxB}} & \small{\textbf{XLM}} & \small{\textbf{AxB + Phon} } \\
        \cmidrule(lr){1-4}
        \small{\bf Mean} & .\small 998 & \small .82 & \small .67  \\
        \small{\bf Variance} & \small 5e-6 & \small .08 & \small .06   \\
        \small{\bf Stdev} & \small .002 & \small .28 & \small .25  \\
        \small{\bf Min} & \small .993 & \small .13 & \small .17  \\
        \bottomrule
    \end{tabular}
    \vspace*{-2mm}
    \caption{Statistics of AxomiyaBERTa (AxB) and XLM's {\tt[CLS]} token cosine similarities compared to those of the pooled output of AxomiyaBERTa with phonological signals (AxB + Phon) over 100 random samples of within-set pairs of Cloze-QA dev set.}
    \label{tab:mean_variance_pan_qa}
    \vspace*{-2mm}
\end{table}

%\textcolor{red} {CAN WE SHORTEN THESE-->}
%This can be a potential use-case for using such phonological signals to train a stronger discriminator between classification decision boundaries, i.e., such articulatory features can help disperse the embedding space for easier classification and thereby, improve task performance. 

To extract phonological features, we used the Assamese grapheme-to-phoneme mapping from \citet{nath-etal-2022-phonetic}, written for the Epitran library \cite{mortensen-etal-2018-epitran}\footnote{This was created using reference resources like Omniglot (\url{https://omniglot.com/writing/assamese.htm}) and Wikiwand/Assamese (\url{https://www.wikiwand.com/en/Help:IPA/Assamese}) along with native-speaker verification.} to convert all text into the International Phonetic Alphabet (IPA). We then used the PanPhon library \cite{mortensen-etal-2016-panphon} to convert the IPA transcriptions into 24 subsegmental features such as place and manner of articulation, voicing, etc.

These feature vectors are padded to the maximum length (across train, test, and dev sets), and then concatenated to either the pooled context embedding (for long-context tasks) or the named-entity token embedding (for NER tasks).
%For multiple-choice tasks, we concatenate them to the pooled embedding outputs of the respective contexts. For NER, we concatenate them to the last hidden state output of the {\it first} subtoken of the named entity. The other subtokens are essentially nullified using zero-padding and using a target label of {\tt -100} for non-named entity subtokens. 
 
\vspace*{-1mm}
\subsubsection{Cloze-style multiple-choice QA}
\label{sssec:cloze-qa}

We fine-tuned AxomiyaBERTa on the Cloze-style Wiki question answering task from the IndicGLUE dataset~\cite{kakwani-etal-2020-indicnlpsuite}. We surrounded both the masked text segment as well as the four candidate answers with the special tokens ({\tt <m>} and {\tt </m>}) and then fed them into the pretrained AxomiyaBERTa model to get pairwise scores with BCE loss. Positive samples were labeled as {\tt 1} and negatives as {\tt 0}. The encoded representation for each sample was a concatenation of the pooled ({\tt[CLS]}) token output, the averaged embedding for the masked text segment ({\tt arg1}), that of the candidate answer ({\tt arg2}), and the element-wise multiplication of {\tt arg1} and {\tt arg2}. This was input into a pairwise scorer {\it a la} \citet{caciularu-etal-2021-cdlm-cross}. We fine-tuned our model (with and without phonological attention) with the pairwise scorer head for 5 iterations with a batch size of 80, a scorer head learning rate of {\tt 1e-4} and a model learning rate of {\tt 2e-5}.

\vspace*{-1mm}
\subsubsection{Named Entity Recognition (NER)}
\label{sssec:ner}

For NER, we fine-tuned and evaluated AxomiyaBERTa on two datasets: WikiNER~\cite{pan-etal-2017-cross} and AsNER~\cite{pathak-etal-2022-asner}. For both datasets, we fed in the tokenized sentence while masking out all sub-word tokens except the first of each word. We used a token-classification head fine-tuned using a multi-class cross-entropy loss for the label set of the respective datasets. For our model without phonological signals, we fine-tuned for 10 epochs with a learning rate of {\tt 2e-5} with a linear LR scheduler and a batch size of 20. For our phonological attention-based model, we fine-tuned for 20 epochs with a batch size of 40 while keeping all other hyperparameters the same.

\vspace*{-1mm}
\subsubsection{Wikipedia Section Title Prediction}
\label{sssec:wiki-titles}

Like Cloze-QA, this task comes from IndicGLUE~\cite{kakwani-etal-2020-indicnlpsuite}.  Fine-tuning for this task was quite similar to that of Cloze-QA, except we did not surround the candidates or the contexts with the trigger tokens.  We fed in the Wikipedia section text and candidate title and optimized the multi-class cross entropy loss with a multiple choice head. We fine-tuned for 20 epochs with a batch size of 40. For the phonologically-aware model, we concatenated the articulatory signals to the pooled embedding output for each sample and fine-tuned our model for 200 iterations with a batch size of 40. We used a smaller model learning rate of {\tt 1e-6} and a classifier head learning rate of {\tt 9.5e-4} for both these models. 
 
\vspace*{-1mm}
\subsubsection{Pairwise Scorer for Assamese CDCR}
\label{sssec:cdcr}

Coreference resolution in a cross-document setting (CDCR) involves identifying and clustering together mentions of the same entity across a set of documents \cite{lu2018event}. Following CDCR approaches in \citet{cattan-etal-2021-cross} and \citet{caciularu-etal-2021-cdlm-cross}, we trained a pairwise scorer with BCE loss over all antecedent spans for each sentence containing an event (across all documents) while ignoring identical pairs. We generated concatenated token representations from Transformer-based LMs by joining the two paired sentences after surrounding the event mentions with the special trigger tokens. These representations were input to the pairwise scorer (PS) to calculate {\it affinity scores} between all those pairs. Mathematically, 

% As far as we know, there has been no such research for CDCR in the case of a very-low resource like Assamese, mostly due to the dearth of annotated corpora. Therefore, we translate the entire ECB+ corpus to Assamese with Microsoft Translator along with a native speaker verification process to weed out and fix mis-translations. 

\vspace*{-2mm}
{\small
$$
Scores(i, j) = PS([CLS], f(x), f(y), f(x) * f(y)),
$$
}where $[CLS]$ represents the pooled output of the entire sentence pair, $f(x)$ and $f(y)$ are the representations of the two events (in context) and $*$ represents element-wise multiplication. 

We trained the Pairwise Scorer for 10 epochs for all baseline models as well as AxomiyaBERTa. At inference, we used a connected-components clustering technique with a tuned threshold to find coreferent links. For baselines and ablation tasks, we calculated coreference scores using a lemma-based heuristic, and fine-tuned four other popular MLLMs using the same hyperparameters. More details and analysis are in Appendix~\ref{app:cdcr}.

% Thereafter, we ablate against a lemma-heuristic baseline along with fine-tuning four other popular MLLMs using the same hyperparameters. See the Appendix..for more details and analyses.

% More details and analysis can be found in the Appendix.

% from the affinity scores and form the system coreference clusters. 

% For baselines and ablation tasks, we calculate coreference scores using a lemma-based string-matching heuristic, and fine-tune four other popular MLLMs using the same hyperparameters. We also include results from phonologically-aware AxomiyaBERTa. More details and analysis can be found in the Appendix.

\vspace*{-2mm}
\section{Evaluation}
\label{sec:eval}
\vspace*{-2mm}

Table~\ref{table:dataset_details} shows the number of samples in the train, dev, and test splits, and the padding length, for all tasks we evaluated on.  For Cloze-QA and Wiki-Titles, we evaluated on IndicGLUE. For NER, we evaluated on AsNER and WikiNER.  For our novel coreference task, we evaluated on the translated ECB+ corpus, where the ratio of coreferent to non-coreferent pairs in the test set is approximately 1:35.  We conducted exhaustive ablations between native and the phonologically-aware models for each task, and compared to previously-published baselines where available.  For Cloze-QA, we created a train/test split of approximately 4.5:1. We fine-tuned off-the-shelf IndicBERT and MBERT on AsNER for 10 epochs on 1 NVIDIA RTX A6000 48 GB device with a batch size of 20.
% For AsNER, we create a training/text split of ... 
\begin{table}[!htb]
\centering
 \begin{tabular}{rllll} 
\toprule
\small{\bf Features} & \small{\bf Train} & \small{\bf Dev} & \small{\bf Test} & \small{\bf Pad-Len} \\
        \cmidrule(lr){1-5}
        \small{{\bf Cloze-QA}}  & \small 8,000 & \small 2,000 & \small 1,768 & \small 360 \\ 
        \small{{\bf Wiki-Titles}} & \small 5,000 & \small 625 & \small 626 & \small 1,848 \\ % Put in the correct numbers!
        \small{{\bf AsNER}} & \small 21,458 & \small 767 & \small 1,798 & \small 744 \\
        \small{{\bf WikiNER}} & \small 1,022 & \small 157 & \small 160 & \small 480 \\
        \small{{\bf T-ECB+}} & \small 3,808 & \small 1,245 & \small 1,780 & \small 552 \\
        % T-ECB+ & 594 & 196 & 206\\
 \bottomrule
 \end{tabular}
 \vspace*{-2mm}
 \caption{Table showing distribution of train/dev/test splits for all tasks. T-ECB+ signifies number of event mentions in the translated ECB+ corpus, keeping special trigger tokens in place. ``Pad-Len'' represents the maximum padded length of the articulatory feature embeddings generated from PanPhon for all three splits.}
\label{table:dataset_details}
\vspace*{-2mm}
\end{table}

\vspace*{-2mm}
\section{Results and Discussion}
\label{sec:results}
\vspace*{-2mm}

Table~\ref{tab:final_results} shows Test F1 Scores/Accuracy for AxomiyaBERTa for the various short-context (classification) and long-context (multiple-choice) tasks. We compared baselines from previous works and newly fine-tuned baselines for certain tasks. We used the same pretrained model for all experiments with task fine-tuning heads consistent with previous benchmarks \cite{kakwani-etal-2020-indicnlpsuite}. One exception is the Cloze-QA task where we dealt with task-specific severe anisotropy with embedding dispersal.

% # I think CDCR should be a different table since it is a novel task and we should surely report the recall, precision and F1 scores from at least three metrics
% like BCUB, MUC and Ceafe

\begin{table*}[h!]
    \centering
    \begin{tabular}{rllll}
        \toprule
         \small{{\bf Models}} & \small{{\bf Cloze-QA}} & \small{{\bf Wiki-Titles}} & \small{{\bf AsNER (F1)}} & \small{{\bf WikiNER (F1)}}  \\
        \cmidrule(lr){1-5}
        \small{XLM-R} & \small{27.11}& \small{56.96} &\small{69.42} &\small{66.67} \\
        \small{MBERT} & \small{29.42}& \small{\bf 73.42} &\small{68.02*} &\small{\bf 92.31} \\
        \small{IndicBERT-{\sc base}} & \small{40.49}& \small{65.82} &\small{68.37*} &\small{41.67} \\
        %\small{IndicBERT-large} & \small{30.03}& \small{56.96} &\small{82.19*} &\small{43.48} & \small{-}\\
        \small{MuRIL} & \small{-}& \small{-} &\small{80.69} &\small{-} \\
        \cdashline{1-5}
        \small{AxomiyaBERTa} & \small{46.66}& \small{26.19} &\small{81.50} &\small{72.78} \\
        \small{AxomiyaBERTa + Phon}  & \small{\bf 47.40} & \small{59.26} &\small{\bf 86.90} & \small{81.71} \\

        %\small{AxomiyaBERTa + Phon (partial overlap)} & \small{-}& \small{-} &\small{85.2} &\small{\underline{72.8}} \\
        % \small{AxomiyaBERTA + EmbDisp} & & & & &\\
        \bottomrule
    \end{tabular}
    \vspace*{-2mm}
    \caption{Test F1 Scores/Accuracy for AxomiyaBERTa on all evaluation tasks, compared to previous baselines and our fine-tuned baselines. ``AxomiyaBERTa + Phon'' shows results for phonologically-aware AxomiyaBERTa. %Partial boundary matches ensuring token-type overlap unlike~\citet{segura-bedmar-etal-2013-semeval} are shown in the last row for both NER tasks.
    AsNER scores with a * represent versions we fine-tuned for this task. For Cloze-QA, Wiki-Titles and WikiNER, other model performances are from~\citet{kakwani-etal-2020-indicnlpsuite}. {\bf Bold} indicates best performance.}
    \label{tab:final_results}
\vspace*{-2mm}
\end{table*}

\vspace*{-1mm}
\subsection{Short-context: AsNER and WikiNER}
\label{ssec:results-short-c}
\vspace*{-1mm}

AxomiyaBERTa achieved SOTA performance on the AsNER task and outperformed most other Transformer-based LMs on WikiNER. %\textcolor{blue}{These  baselines include Indian-language (IN) specific LMs like IndicBERT-base that was specifically trained on IN-language parallel corpora as well as much larger models like XLM (with 570 M parameters).} 

\begin{figure}
  \centering
  \includegraphics[width=0.235\textwidth,trim={77px 6px 92px 27px},clip]{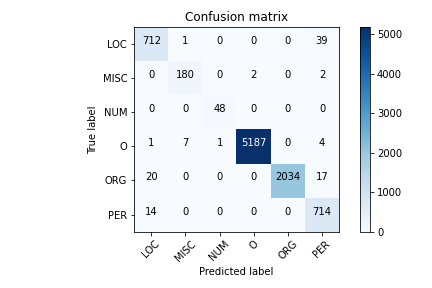}  \includegraphics[width=0.235\textwidth,trim={77px 6px 92px 27px},clip]{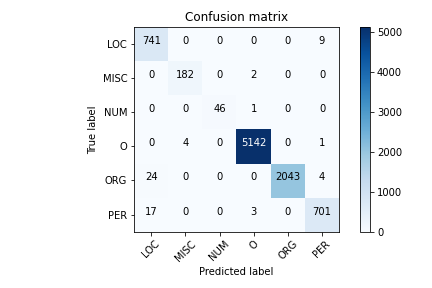} \\
  \includegraphics[width=0.235\textwidth,trim={77px 11px 92px 22px},clip]{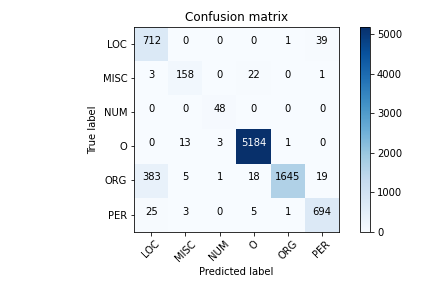}  \includegraphics[width=0.235\textwidth,trim={77px 11px 92px 22px},clip]{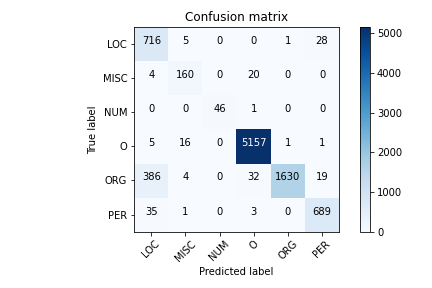} \\
  \vspace*{-2mm}
\caption{Top: Confusion matrices showing AxomiyaBERTa performance on AsNER without [L] and with [R] phonological awareness. Bottom: IndicBERT [L] and MBERT [R] performance on AsNER.} 
\label{fig:confusion_matrix}
\vspace*{-4mm}
\end{figure}

\vspace*{-2mm}
\paragraph{Phonologically-aware AxomiyaBERTa}
Our experiments suggest that phonological signals are informative additional features for short-context tasks like NER for low-resourced, smaller models like AxomiyaBERTa. Table~\ref{tab:final_results} shows that phonologically-aware AxomiyaBERTa outperformed non-phonological (hereafter ``native'') AxomiyaBERTa by >5 F1 points on AsNER, with an even greater improvement (10 F1 points) on WikiNER. AxomiyaBERTa also outperformed other baselines for both tasks, with the exception of MBERT on Wiki-based tasks.\footnote{Wikipedia comprises almost all of MBERT's training data. MBERT does not support Assamese, but does support Bengali, and Assamese is written using a variant of the same script. Named entities are often written identically in Bengali and Assamese, which could explain this trend.} Fig.~\ref{fig:confusion_matrix} shows confusion matrices of performance on AsNER.

IndicBERT and MBERT misclassified \textit{ORG} tokens as \textit{LOC} 16 times as much as AxomiyaBERTa. Specific cases include sub-tokens like {\bng inUyr/k} (\textipa{/niujO\textturnr k/}, ``New York'') or {\bng ichNNGgapurW} (\textipa{/siNgapu\textturnr/}, ``Singapore''), that are actually parts of entities like {\bng E-Sh/TarW \underline{ichNNGgapurW}} (\textipa{/e sta\textturnr~siNgapu\textturnr/}, ``A-Star Singapore'') or {\bng \underline{inUyr/k} b/laD ecN/TarW} (\textipa{/niujO\textturnr k blad senta\textturnr/}, ``New York Blood Center''). This suggests that smaller, monolingual models like AxomiyaBERTa with a reduced sequence length and no NSP training objective are optimized for NE classification tasks with greater attention to local context (since the average sentence containing NEs is $\sim$6 tokens). 

Better overall performance on AsNER than on WikiNER can be partially attributed to having one fewer class and a more balanced distribution between categories. AsNER performance likely benefited from a greater phonological signal and more data to tune on (Table~\ref{table:dataset_details}) whereas WikiNER text samples are, on average, longer than 128 tokens (AxomiyaBERTa's maximum token length) possibly causing a performance loss due to truncated context.

% . WikiNER’s token length is >128 (pretrained AxomiyaBERTa’s maxlen). 

% While richer phonological signal increases performance within this truncated context (Table 4: 26\% -> 59\%), it still lags larger models.}

%than larger models that generally capture long-range semantic dependencies better. On the other hand, models like MBERT with larger vocabularies and training data might have seen more samples like {\beng নিউয়র্ক} within a context that \textit{implies} LOC than our AxomiyaBERTa training corpora.

\vspace*{-2mm}
\paragraph{Phonological Signals: A Disambiguation Tool}
Even though phonologically-aware AxomiyaBERTa took a hit on identifying {\it O} tokens, it compensated with improved results across other classes. Phonologically-aware AxomiyaBERTa also reduced misclassifications of {\it ORG} tokens as {\it PER} compared to all other models, including native AxomiyaBERTa. Specific cases include tokens that imply persons, e.g., {\bng sWamiinathn} or {\bng saHa}, but are actually part of {\it ORG} NEs, e.g., {\bng  \underline{sWamiinathn} kimchn} (``Swaminathan Commission'') or {\bng \underline{saHa} IniSh/TiTUT Aph iphijk/s} (``Saha Institute of Physics''). Similarly, in the WikiNER task, phonological attention reduced misclassification of {\it B-ORG} and {\it I-ORG} tokens as {\it B-PER} and {\it I-PER} respectively (see Appendix~\ref{app:short-c}). These results suggest phonological inputs help enrich embeddings of smaller-sized LMs to distinguish such ambiguous tokens.

\vspace*{-1mm}
\subsection{Long-context: Multiple Choice}
%\vspace*{-1mm}

On Wiki-Titles, phonological AxomiyaBERTa does better with semantically harder multiple-choice sets, which have a higher average cosine similarity between the candidate options. Native AxomiyaBERTa fails on these samples. As shown in Table~\ref{table:mchoice_analysis}, {\bf P+N-} has the highest average cosine similarity between the sample sets, suggesting that there are cases where phonological signals compensate for low semantic variation among candidate options. On the other hand, native AxomiyaBERTa tends to do better with multiple-choice sets that have wider (relative) semantic variation within that set, on average. Since the overall distribution of embeddings in this task is still extremely close, this suggests that phonological signals are doing for Wiki-Titles what the embedding disperser did for Cloze-QA (see Sec.~\ref{sssec:disperser}).

\begin{table}[!htb]
\vspace*{-2mm}
\centering
 \begin{tabular}{rllll} 
\toprule
\small{\bf } & \small{\bf P+N-} & \small{\bf P-N+} & \small{\bf P+N+} & \small{\bf P-N-} \\
        \cmidrule(lr){1-5}
        \small{\bf Cos-sim} & \small .98844 & \small .98829 & \small .98824 & \small .98838 \\ 
 
 \bottomrule
 \end{tabular}
 \vspace*{-2mm}
 \caption{Average cosine similarities between within-set samples on the Wiki-Titles test set for native (N) and phonological (P) AxomiyaBERTa. ``+'' and ``-'' represent correct and incorrect samples respectively, e.g., {\bf P+N-} shows samples phonological AxomiyaBERTa answered correctly that the native variant did not.}
\label{table:mchoice_analysis}
\vspace*{-4mm}
\end{table}

\vspace*{-1mm}
\subsection{Novel Task: Event Coreference on Translated ECB+}
\vspace*{-1mm}

%We might have to make two tables here since it look too small !

%We might have to make two tables here since it look too small !

\begin{table*}[h!]
    \centering
    
 \resizebox{\textwidth}{!}{%
\begin{tabular}{rccccccccccccccccccc} 
\toprule
 
%\multirow{2}[1]{*}{CDCR Models}   \\

% \cmidrule(lr){2-11}  
\multirow{2}[2]{*}{CDCR Models} & \multicolumn{3}{c}{BCUB} & \multicolumn{3}{c}{MUC}  & \multicolumn{3}{c}{CEAF-e}  & \multicolumn{3}{c}{BLANC} & C-F1    \\
\cmidrule(lr){2-4} \cmidrule(lr){5-7} \cmidrule(lr){8-10} \cmidrule(lr){11-13}  \cmidrule(lr){14-14} 
 & P & R & F1 & P & R & F1 & P & R & F1 &  P & R & F1 & \\

\tt \textbf{Lemma Baseline} & 75.81 & 60.24 & 67.14 & 64.59 & 54.25 & 58.97 & 61.36 & 73.25 & \textbf{66.78} & \textbf{74.97} & 60.40 & \textbf{64.66} & \textbf{64.29} \\
\cdashline{1-14}
\tt \textbf{XLM-100}$^\dagger$ & 5.31 & \textbf{97.55} & 10.08 & 54.17 & \textbf{97.84} & 69.73 & 30.99 & 0.73 & 1.42 & 49.78 & 50.00 & 49.89 & 27.07  \\
\tt \textbf{IndicBERT-{\sc base}} & 74.48 & 51.93 & 61.19 & 44.03 & 21.94 & 29.29 & 40.80 & 65.59 & 50.31 & 52.09 & 55.41 & 52.93 & 46.93 \\
\tt \textbf{MuRIL} & \textbf{93.53} & 48.33 & 63.73 & \textbf{68.18} & 9.23 & 16.26 & 41.56 & \textbf{85.09} & 55.85 & 54.78 & 53.31 & 53.91 & 45.28 \\
\cdashline{1-14}
\tt \textbf{AxomiyaBERTa} & 34.68 & 85.98 & 49.42 & 62.40 & 80.51 & \textbf{70.30} & \textbf{67.63} & 43.85 & 53.20 & 53.00 & \textbf{87.75} & 54.23 & 57.64 \\
\tt \textbf{AxomiyaBERTa + Phon} & 70.00 & 64.58 & \textbf{67.18} & 64.11 & 44.71 & 52.68 & 50.18 & 68.57 & 50.18 & 56.22 & 68.65 & 59.19 & 59.27 \\
% \tt \textbf{Lemma Baseline (Ours)} & 42.63 & 75.52 & 55.4 & 71.25 & 81.12 & 75.86 & 56.24 & 40.25 & 46.92 & 30.65 & 40.25 & 46.92 & 30.65 \\
 %\addlinespace[0.5cm]  
 
%\midrule  

\bottomrule
\end{tabular}}
\vspace*{-2mm}
\caption{Event coreference results on Assamese (translated) ECB+ test set from pairwise scorer using AxomiyaBERTa, compared with other Transformer-based LMs and the lemma-based heuristic. {\bf Bold} indicates best overall performance per metric. ``C-F1'' is CoNLL F1. $^\dagger$We evaluate XLM-100 to compare performance on this task of a slightly larger model than XLM-R where most other major design factors remain the same.}
\label{fig:cdcr_events_fulltable}
\vspace*{-2mm}
\end{table*}

Table~\ref{fig:cdcr_events_fulltable} shows event coreference resolution results on the translated ECB+ test set using a tuned affinity-threshold ($T$). These results include both within- and cross-document system outputs from AxomiyaBERTa, other Transformer-based LMs, and a lemma-based heuristic.\footnote{We found an affinity threshold ($T=7$) to work for all models except phonologically-aware AxomiyaBERTa ($T=0$) and XLM-100 ($T=-1.94$). For the latter, we use the mean of all scores due to an extremely narrow distribution as shown in the Appendix. More analysis of why this happens is the subject of future work.}

AxomiyaBERTa often outperformed the lemma-similarity baseline and other LMs. Native and phonological AxomiyaBERTa have the best MUC and BCUB F1 scores, respectively, while also outperforming all other Transformer-based LMs on BLANC and CoNLL F1. Phonologically-aware AxomiyaBERTa also outperforms native AxomiyaBERTa by almost 2 F1 points on CoNLL F1. More importantly, the phonological signals help detect more challenging coreferent links where mere surface-level lemma similarity can fail.
% True Positive: coreferent/positive mention pair with high surface similarity.
% False Negative: coreferent/positive mention pair with low surface similarity.
While native and phonological AxomiyaBERTa performed comparably, the true positives retrieved by the phonological version contained a higher proportion of non-similar lemmas, which were usually missed by the lemma heuristic.  Meanwhile, native AxomiyaBERTa retrieved results with more similar lemmas, labeling more non-similar lemma pairs as false negatives
%I.e., phonological AxomiyaBERTa has a higher (lower) percentage of non-overlapping (different) lemma pairs in its true positive (false-negative)  distribution on the T-ECB+ test set compared to baseline AxomiyaBERTa.
(Table~\ref{table:cdcr_analysis}). Compared to the other Transformer models, this also had the effect of increasing precision according to most metrics, though at the cost of decreasing recall.  However, the increased precision was usually enough to increase F1 overall, pointing to the utility of phonological signals in detecting more challenging cases. We hypothesize that this is because these challenging pairs may consist of synonyms and/or loanwords, and phonological signals helped correlate these different surface forms, which in addition to the semantic information at the embedding level helps create coreference links. 

For instance, {\bng kncail/tNNG} (\textipa{/kOnsaltiN/}, ``consulting'') and {\bng IiNJ/jinyairWNNG} (\textipa{/indZinija\textturnr iN/}, ``engineering'') denote two coreferent events pertaining to the same company (EYP Mission Critical Facilities). Since both are borrowed words that maintain the original phonological form, phonological signals can help pick out unique articulation beyond surface-level lemma similarity. Similarly, in cases of synonyms like {\bng mrRtuYrW} (\textipa{/m\textturnr ittu\textturnr/}, ``(of) death'') and {\bng HtYa} (\textipa{/HOtta/}, ``killing''), which do not share surface-level similarity yet are coreferent, phonological signals can help. Where lemmas are already similar, phonological signals provide little extra information.

We should note that for coreference, the specific metric used matters a lot. For instance, almost 33\% of the ECB+ dataset across all three splits consists of singleton mentions. Since MUC score is not as sensitive to the presence of singletons as BCUB~\cite{kubler2011singletons}, this could explain AxomiyaBERTa's (and XLM's) relative drop in performance on the BCUB metric. On the other hand, the lower CEAF-e F1 score may be due to CEAF-e's alignment algorithm, which tends to ignore correct coreference decisions when response entities are misaligned~\cite{moosavi-strube-2016-coreference}.

Ablations between native and phonological AxomiyaBERTa showed that where lemmas for a pair of potentially coreferent events are identical (e.g., {\bng AarWm/bh} - \textipa{/a\textturnr Omb\super Ho/}, ``start''), non-phonological representations primarily determine the pairwise scores and the coreference decision.  Table~\ref{table:cdcr_analysis} shows that even though phonological signals tend to disambiguate harder event pairs, decreased performance (e.g., MUC F1 phonological vs. native AxomiyaBERTa) could be due to native representations of the same-lemma pair being weakly correlated with the pairwise scores, a possibility when a coreferent event pair has high contextual dissimilarity. Phonological signals may add noise here.

We also see that the lemma-based heuristic baseline is overall a very good performer.  While this may be a property of the nature of coreference tasks in general or specific to a dataset (as a high percentage of coreferent events use the same lemma), we must also allow for the possibility that this may also be an artifact of translation noise.  Since we used an automatically-translated version of the ECB+ corpus (albeit with some native speaker verification), and since Assamese is still a low-resource language, the decoder vocabulary of the translator may be limited, meaning that synonymous different-lemma pairs in the original corpus may well have been collapsed into same-lemma pairs in the translation, artificially raising the performance of the lemma heuristic.

\begin{table}[!htb]
 \vspace*{-2mm}
\centering
 \begin{tabular}{rllll} 
\toprule
\small{\bf Models} & \small{\bf TP} & \small{\bf L1} & \small{\bf L2} & \small{\bf Diff-Rate} \\
        \cmidrule(lr){1-5}
        \small{{\bf XLM-100}}  & \small 6,361 & \small 1,441 & \small 4,920 & \small .773 \\ 
        \small{{\bf IndicBERT}} & \small 101 & \small 46 & \small 55 & \small .545 \\ % Put in the correct numbers!
        \small{{\bf MuRIL}} & \small 62 & \small 21 & \small 41 & \small .661 \\
        \small{{\bf AxB}} & \small 1,833 & \small 466 & \small 1,367 & \small .746 (.98)\\
        \small{{\bf AxB + Phon}} & \small 956 & \small 81 & \small 875 & \small .915 (.93)\\
        % T-ECB+ & 594 & 196 & 206\\
 \bottomrule
 \end{tabular}
 \vspace*{-2mm}
 \caption{Distribution of same (L1) and different (L2) event lemma samples in the true positive (TP) distribution of the T-ECB+ test set. ``Diff-Rate'' is the percentage of different lemma samples within TPs ($= L2/TP$). Values in parentheses show the equivalent distribution within false negatives for comparison.}
\label{table:cdcr_analysis}
\vspace*{-4mm}
\end{table}

\vspace*{-2mm}
\section{Conclusion and Future Work}
\label{sec:conc}
\vspace*{-2mm}

In this paper, we presented a novel Transformer model for Assamese that optionally includes phonological signals. We evaluated on multiple tasks using novel training techniques and have demonstrated SOTA or comparable results, showing that phonological signals can be leveraged for greater performance and disambiguation for a low-resourced language. AxomiyaBERTa achieves SOTA performance on short-context tasks like AsNER and long-context tasks like Cloze-QA while also outperforming most other Transformer-based LMs on WikiNER, with additional improvement resulting from the phonologically-aware model. For challenging tasks like CDCR, we have shown that both AxomiyaBERTa outperformed other Transformer-based LMs on popular metrics like BCUB, MUC, and CoNLL F1.

More generally, we have shown that strategic techniques for optimizing the embedding space and language-specific features like phonological information can lower the barrier to entry for training language models for LRLs, making it more feasible than before with lower amounts of data and a ceiling on compute power. Our experiments suggest phonological awareness boosts performance on many tasks in low-resource settings. Future models for other LRLs can leverage our ideas to train or fine-tune their own models. Since smaller models tend toward anisotropy, embedding dispersal may pave the way for more such performant LRL models.

Future work may include incorporating phonological signals during pretraining instead of fine-tuning, carrying out evaluations against semantically harder tasks like paraphrasing or emotion detection, zero-shot transfer to similar languages, and a contrastive learning framework with a triplet loss objective for CDCR.

Our trained checkpoints are available on HuggingFace at \url{https://huggingface.co/Abhijnan/AxomiyaBERTa}. We hope this resource will accelerate NLP research for encoding language-specific properties in LRLs.

% These findings should motivate more research into phonological-aware language models in resource-constrained settings. 

% Expanded FW section for final copy
%An obvious experiment as future work is to explicitly incorporate phonological signals during the model pretraining phase to enrich the embedding space with such signals at/from the source. While this might come at an additional cost of computation, this would allow us to understand and interpret how and to what extent encoder-only models like BERT can encode language-specific phonology. Another direction can be to carry out a more exhaustive evaluation on self-curated semantically harder tasks like paraphrase detection or emotion analysis. One could also experiment with zero-shot transfer using the phonological AxomiyaBERTa model, especially for geographically proximate, tonal Tibeto-Burman languages like Mishing and Bodo which are even more low-resourced but share crucial linguistic similarities with Assamese. From a task-specific perspective, researchers might benefit from a reduced translation noise in the ECB+ dataset with more human supervision, a direction we are committed to. There is also a case to be made for using a contrastive learning framework with a triplet loss objective for CDCR as such methods are gaining popularity in English CDCR. We hope the availability of this resource will accelerate NLP research for encoding other crucial linguistic properties like morphology, tonality etc. in LRLs.

%\section*{Acknowledgements}

\section*{Limitations}

Let us begin with the obvious limitation: AxomiyaBERTa only works on Assamese. In addition, since Assamese comprises a number of dialects and we trained on internet-sourced data, we have no clear evidence regarding which dialects AxomiyaBERTa is most suited to or if it performs as well on non-standard dialects.

AxomiyaBERTa did not perform all that well on Wikipedia Title Selection, compared to other Transformer-based models. Our best result is on par with XLM-R and close to IndicBERT-{\sc base}, but well below MBERT performance.  We hypothesize that the amount of Wikipedia training data in MBERT is a cause of this, but we find that phonological attention makes a big difference in AxomiyaBERTa's performance (increasing accuracy from 26\% to 59\%).  Nonetheless, the reasons behind this subpar performance, and whether AxomiyaBERTa can be improved for this task without, say, overfitting to Wikipedia, need further investigation.

\section*{Ethics Statement}

\paragraph{Data Usage}  Because of the publicly-available, internet-sourced nature of our training data, we cannot definitively state that the current version of AxomiyaBERTa is free of bias, both in terms of outputs nor, as mentioned in the limitations section, if there are dialect-level biases toward or against certain varieties of Assamese that may be trained into the model.  Such investigations are the topic of future research.

\paragraph{Resource Usage and Environmental Impact}  At 66M parameters, AxomiyaBERTa is a smaller language model that is relatively quick to train and run.  Training was conducted on single GPU devices.  Pretraining AxomiyaBERTa took approximately 3 days, and task-level fine-tuning took roughly 30 minutes for non-phonological AxomiyaBERTa  and 1-2 hours for phonological AxomiyaBERTa (depending on the task).  Training the pairwise scorer for CDCR took 12-19 minutes.  Training and fine-tuning took place on the same hardware.  For comparison, fine-tuning IndicBERT and MBERT on the AsNER dataset for evaluation took roughly 20-30 minutes each.  These figures indicate that relative to work on other Transformer models, training and evaluating AxomiyaBERTa (including running other baselines for comparison) comes with a comparatively lower resource usage and concomitant environmental impact.  This lower resource usage also has implications for the ``democratization'' of NLP, in that we have demonstrated ways to train a performant model with fewer local resources, meaning less reliance on large infrastructures available to only the biggest corporations and universities. 

\paragraph{Human Subjects}  This research did not involve human subjects.

\section*{Acknowledgments}

We would like to thank the anonymous reviewers whose feedback helped improve the final copy of this paper. Special thanks to Ibrahim Khebour for helping with the phonological feature extraction process for the Wikipedia Section Title Prediction task.

%Scientific work published at ACL 2023 must comply with the ACL Ethics Policy.\footnote{\url{https://www.aclweb.org/portal/content/acl-code-ethics}} We encourage all authors to include an explicit ethics statement on the broader impact of the work, or other ethical considerations after the conclusion but before the references. The ethics statement will not count toward the page limit (8 pages for long, 4 pages for short papers).

% Entries for the entire Anthology, followed by custom entries
\bibliography{anthology,custom, nath_thesis}
\bibliographystyle{acl_natbib}

\appendix

\section{Training Configuration}
\label{app:train-config}

Table~\ref{tab:axberta-config} shows the pretraining configuration for AxomiyaBERTa.
\begin{table*}[h!]
    \centering
    \begin{tabular}{llr}
        \toprule
        \small {\bf Parameters} & \small {\bf Config }  \\
        \cmidrule(lr){1-2}
      \small \texttt{architecture} & \small AlbertForMaskedLM \\
      \small \texttt{attention\_probs\_dropout\_prob} & \small 0.1  \\
      \small \texttt{bos\_token\_id} & \small 2  \\
      \small \texttt{classifier\_dropout\_prob} & \small 0.1  \\
      \small \texttt{embedding\_size} & \small 128  \\
      \small \texttt{eos\_token\_id} & \small 3  \\
      \small \texttt{hidden\_act} & \small {\tt gelu}  \\
      \small \texttt{hidden\_dropout\_prob} & \small 0.1  \\
      \small \texttt{hidden\_size} & \small 768  \\
      \small \texttt{initializer\_range} & \small 0.02  \\
      \small \texttt{inner\_group\_num} & \small 1  \\
      \small \texttt{intermediate\_size} & \small 3072  \\
      \small \texttt{layer\_norm\_eps} & \small 1e-05  \\
      \small \texttt{max\_position\_embeddings} & \small 514  \\
      \small \texttt{num\_attention\_heads} & \small 12  \\
      \small \texttt{num\_hidden\_groups} & \small 1  \\
      \small \texttt{num\_hidden\_layers} & \small 6  \\
      \small \texttt{position\_embedding\_type} & \small ``absolute"  \\
      \small \texttt{transformers\_version} & \small ``4.18.0"  \\
      \small \texttt{vocab\_size} & \small 32001  \\
        \bottomrule
    \end{tabular}
    \vspace*{-2mm}
    \caption{AxomiyaBERTa Model configuration trained on a monolingual Assamese corpus.}
    \label{tab:axberta-config}
    \vspace*{-4mm}
\end{table*}

\section{Further Details on Embedding Disperser}
\label{app:emb-disp}

\begin{figure*} 
    \centering
    \includegraphics[scale=1.9,width=\textwidth]{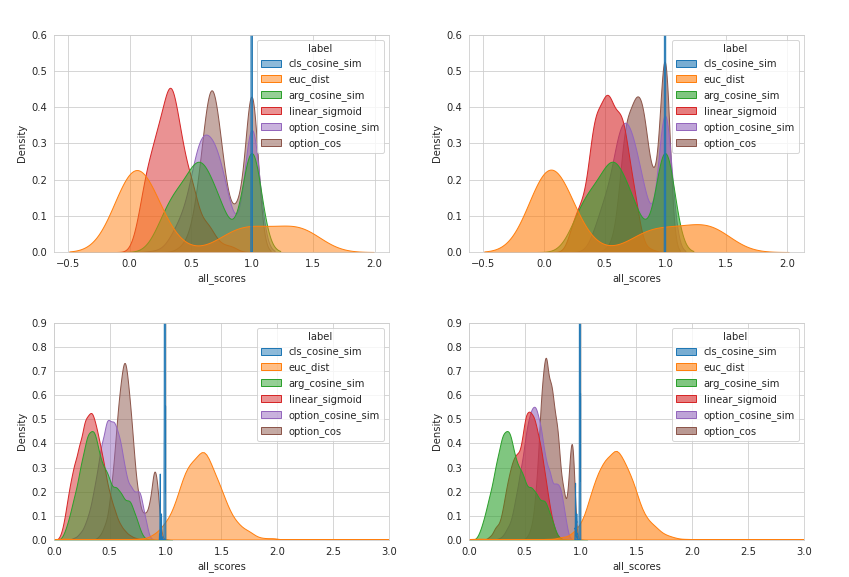} 
    \vspace*{-2mm}
    \caption{Kernel density estimation plots from various feature sets of the embedding disperser model. The figure on the top left represents ``within set'': cosine similarities between each set of candidate answers plus context and the remaining three pairs in that set, e.g., if $Q$ is the question/context and $A/B/C/D$ are the candidate answers, $S_C(Q+A, Q+i)$, where $i$ represents one of the remaining three candidates $B$, $C$, $D$. The figure on the bottom left represents ``beyond-set'' cosine similarities: all pairs in a candidate-plus-answers set are compared to other such sets with the cosine similarity metric. We run our experiments for 100 randomly selected sets from the Cloze-QA dev set. The top right (within-set) and bottom right (beyond-set) figures are equivalent figures for our models with phonological awareness.}
    \label{fig:cosine_embed}
    \vspace*{-4mm}
\end{figure*}

Fig.~\ref{fig:cosine_embed} shows KDE plots for outputs of different components of the embedding disperser, showing the contrast between features within-set and beyond-set for Cloze-QA samples, and showing the difference between AxomiyaBERTa with phonological awareness and without. The {\tt option\_cos} label (brown) shows an interesting phenomenon. This is the output of the embedding disperser at inference (Auxiliary Discriminator in Fig.~\ref{fig:emb_disperser}) and represents a 128-dimensional embedding output from the {\tt [CLS]} token concatenated with {\tt arg2} or the candidate answer input. We see a distinct shift in cosine similarity scores between within-set and beyond-set with one peak very close to 1 in the case of the within-set pairs while getting clearly dispersed to a lower cosine similarity score in the case of beyond-set pairs. This phenomenon is even further accentuated by feeding phonological signals to the disperser. In this case, as shown in the top right plot, the cosine similarity peak for {\tt option\_cos} has a much higher density compared to the non-phonological disperser while the overall distribution is shifted to a higher cosine similarity. 

Another interesting trend is the {\tt linear\_sigmoid} label (red) which is the sigmoidal output of the linear layer of the disperser, trained with a combination of cosine embedding loss and BCE loss when fed an input of the combined {\tt arg1} and {\tt arg2} representations generated with the special trigger tokens. In this case, feeding phonological signals to the model reduces dispersion (an inverse trend) in the cosine similarities between within-set and beyond-set pairs (as seen in the top-left plot where this label has a narrower top with a wider bottom). However, this reverse effect is less pronounced than that seen in the {\tt option\_cos} cosine similarity plot, perhaps due to richer contextual information carried by the trigger token representations (the inputs to this layer). In other words, and as shown in the {\tt arg\_cosine\_sim} plot, its dispersion between the within- and beyond-set pairs suggests why such an effect is less-pronounced.

% \begin{figure*} 
%     \centering
%     \includegraphics[scale=1.9,width=\textwidth]{plots/axbert_embeddingdisperser_large.png} 
%     \vspace*{-2mm}
%     \caption{Embedding Disperser Architecture using Cosine Embedding Loss and Binary Cross Entropy (BCE) Loss.}
%     \label{fig:emb_disperser}
%     \vspace*{-4mm}
% \end{figure*}

% the parts of this that don't fit in the main paper go here
Works such as \citet{cai2021isotropy} present evidence of such global token anisotropy in other BERT and GPT-model variants while also suggesting ways to locate/create local isotropic spaces more susceptible for NLP tasks. Interestingly, cosine similarities of output embeddings from our Auxiliary Discriminator ({\tt option\_cos } in Fig.~\ref{fig:cosine_embed_small}) show a marked difference in the extent of anisotropy between within-set and beyond-set pairs, a phenomenon further accentuated with additional phonological signals (top right plot in Fig.~\ref{fig:cosine_embed}). These experiments suggest that a combination of our embedding disperser architecture together with phonological signals (Sec.~\ref{sssec:articulatory} for more details) can effect a shift towards local spaces of isotropy in the embedding space of the fine-tuned AxomiyaBERTa model for Cloze-QA and potentially other tasks. 

\section{Further Discussion on Short-Context Results}
\label{app:short-c}

\begin{figure}
  \centering
  \includegraphics[width=0.235\textwidth,trim={77px 6px 92px 27px},clip]{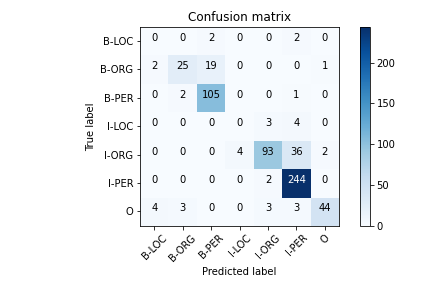}  \includegraphics[width=0.235\textwidth,trim={77px 6px 92px 27px},clip]{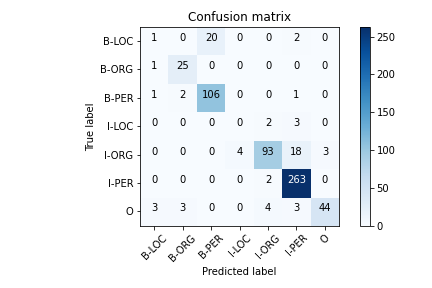}
  \vspace*{-2mm}
\caption{Confusion matrices showing AxomiyaBERTa performance on WikiNER without [L] and with [R] phonological awareness.} 
\label{fig:cosine_embed_wikiner}
\vspace*{-4mm}
\end{figure}

Fig.~\ref{fig:cosine_embed_wikiner} shows native and phonological AxomiyaBERTa performance on WikiNER.  We see comparative performance, but with phonological signals there are fewer confusions of {\it B-ORG} with {\it B-PER} and {\it I-ORG} with {\it I-PER}.  Specific examples are similar to those seen in Sec.~\ref{ssec:results-short-c}, e.g., {\bng sWamiinathn (kimchn)} (``Swaminathan [Commission]'') or {\bng saHa (IniSh/TiTUT Aph iphijk/s)} (``Saha [Institute of Physics]'').  Being organizations named after people, this is a case where phonological signals actually help.  Interestingly, phonological signals also help with NER even when the NEs are broken down into BIO chunks, which was not the case in AsNER.  We should observe that with phonological signals, there is an {\it increase} in {\it B-LOC} tokens classified as {\it B-PER} tokens, which is the topic of future investigation.

\section{Further Discussion on Pairwise Scorer for CDCR on Assamese ECB+}
\label{app:cdcr}

The lemma-based heuristic comes from the fact that a large proportion of coreferent mention pairs can be identified simply because they use the same lemma. These ``easy'' cases gives coreference a very high baseline even when this naive heuristic is used. The long tail of ``harder'' pairs require more sophisticated approaches~\cite{ahmed2023better}.

\begin{figure*} 
    \centering
    \includegraphics[scale=1.9,width=\textwidth]{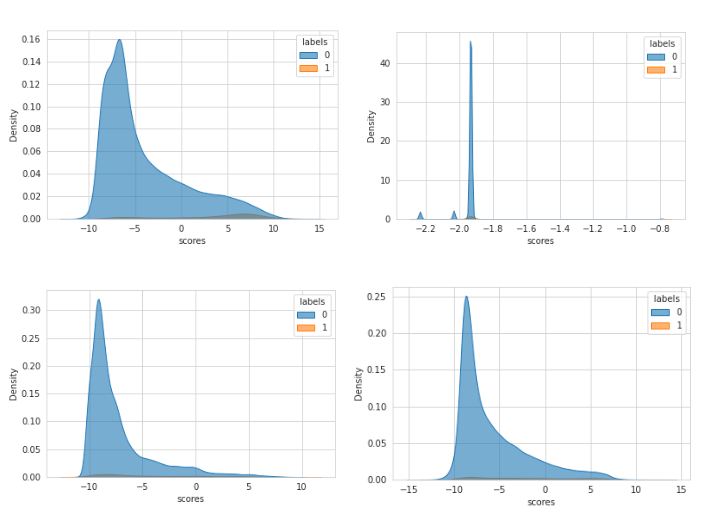} 
    \vspace*{-2mm}
    \caption{Kernel density estimation plots of affinity scores from the pairwise scorer for native AxomiyaBERTa compared to baselines from other Transformer-based LMs.}
    \label{fig:pairwise_scores}
    \vspace*{-4mm}
\end{figure*}

Fig.~\ref{fig:pairwise_scores} shows the affinity scores from the pairwise scorer using various model outputs. AxomiyaBERTa is shown in the top left, followed by (left-to-right, top-to-bottom) XLM-100, MuRIL, and IndicBERT.  We see that AxomiyaBERTa clearly has a more defined separation between the labels, with positive/coreferent samples having higher affinity scores (accounting for the imbalanced distribution of coreferent vs. non-coreferent pairs) compared to the other models.  In particular, XLM-100 shows almost identical ranges of scores for coreferent and non-coreferent pairs, with the only significant difference being the number of each kind of sample, which results in the spike around $T=-1.94$ (cf. Sec.~\ref{sssec:cdcr}).

\end{document}